\documentclass[journal]{IEEEtran}
\usepackage[english]{babel}
\usepackage{nicematrix}
\usepackage[top=2cm,bottom=2cm,left=1.5cm,right=1.5cm,marginparwidth=1.75cm]{geometry}

\usepackage{graphicx,subfigure,mhchem,amsmath,cite}
\usepackage[colorlinks=true, allcolors=blue]{hyperref}

\title{On mesoscale thermal dynamics \\of para- and ortho- isomers of water}
\author{Serge Kernbach \\[3mm]
\small CYBRES GmbH, Research Center of Advanced Robotics and Environmental Science,\\
\small Melunerstr. 40, 70569 Stuttgart, Germany, {\it serge.kernbach@cybertronica.de.com}
\vspace{-6mm}
}

\begin{document}
\maketitle
\thispagestyle{empty}

\begin{abstract}
This work describes experiments on thermal dynamics of pure \ce{H_2O} excited by hydrodynamic cavitation, which has been reported to facilitate the spin conversion of para- and ortho-isomers at water interfaces. Previous measurements by NMR and capillary methods of excited samples demonstrated changes of proton density by 12-15\%, the surface tension up to 15.7\%, which can be attributed to a non-equilibrium para-/ortho- ratio. Beside these changes, we also expect a variation of heat capacity. Experiments use a differential calorimetric approach with two devices: one with an active thermostat for diathermic measurements, another is fully passive for long-term measurements. Samples after excitation are degassed at -0.09MPa and thermally equalized in a water bath. Conducted attempts demonstrated changes in the heat capacity of experimental samples by 4.17\%--5.72\% measured in the transient dynamics within 60 min after excitation, which decreases to 2.08\% in the steady-state dynamics 90-120 min after excitation. Additionally, we observed occurrence of thermal fluctuations at the level of 10$^{-3}$ $^\circ C$ relative temperature on 20-40 min mesoscale dynamics and a long-term increase of such fluctuations in experimental samples. Obtained results are reproducible in both devices and are supported by previously published outcomes on four-photon scattering spectra in the range from -1.5 to 1.5 cm$^{-1}$ and electrochemical reactivity in \ce{CO_2} and \ce{H_2O_2} pathways. Based on these results, we propose a hypothesis about ongoing spin conversion process on mesoscopic scales under weak influx of energy caused by thermal, EM or geomagnetic factors; this enables explaining electrochemical and thermal anomalies observed in long-term measurements.
\end{abstract}


\section{Introduction}

Different ionic reactivity of \ce{H_2O} is demonstrated for several reactions \cite{Kilaj18,kernbach2022electrochemical} that can be related to nuclear spin isomers of water \cite{Tikhonov02}. As pointed out by multiple publications \cite{Pershin14,POULOSE2023814,Pershin22}, spin conversion of para- and ortho- isomers affects not only chemical but also physical properties such as surface tension and capillary effects, hydrodynamic viscosity, evaporation rate and heat capacity.

Such variable physical properties are important to understand dynamics of aqueous solutions in a variety of chemical and biological systems: water movement in a microcapillary system of biological organisms \cite{Pershin09}, persistent thermal effects in neurohumoral regulative system \cite{Kozhevnikov13,kernbach2023Neurocognitive}. Performing laboratory measurements with electrochemical impedance spectroscopy (EIS), we noted electrochemical and thermal anomalies in pure \ce{H_2O}. Such effects differ from both the macro-scale dynamics described by kinetics of chemical reactions (e.g. dissolving of gases) and thermal dependencies \cite{doi:10.1021/ac00140a003}, and the micro-scale dynamics of molecular and thermodynamic phenomena. Typically they appear on meso-scales (30-90 min); their phenomenology and theoretical explanations are widely discussed in the community \cite{KernbachPh2022,NILSSON20122,WANG2022119719}.

A hypothesis is expressed towards the spin conversion processes that potentially operate on mesoscopic scales in fluidic phase \cite{Miani04,PhysRevA.104.052816}; among others, the ice-like structures at water interfaces \cite{PershinIce10,PhysRevLett.101.036101,Wei01,en8099383}, especially at the air-water and metal-water interfaces \cite{Monserrat20,Odendahl22}, are expected to delay achieving of equilibrium state and facilitate the spin conversion. This hypothesis has already a number of experimental confirmations: direct measurements of optical spectra \cite{Pershin09Temp,Pershin14} and NMR spectroscopy \cite{Pershin09NMR}; indirect measurements of ionic reactivity \cite{kernbach2022electrochemical} and evaporation rate \cite{POULOSE2023814,Novikov10}. Discussions on non-equilibrium phase transitions at 4$^\circ C$ \cite{Pershin11JETP} and aquaporin channels \cite{Murata00} contributes to a deeper understanding of this phenomenon.

This work explores thermal effects of suggested spin conversion processes on mesoscales caused by different heat capacity of water isomers. Most of publications are devoted to the heat capacity of isomers of hydrogen \cite{PETITPAS20146533, FUKUTANI2013279}, only several papers deal with the heat capacity of spin isomers of water \cite{Velikov06, Suzuki19}. Authors in \cite{Pershin09NMR,Pershin22} suggest that hydrodynamic cavitation changes the ratio of para- and ortho- isomers by 12\%-15\%; we replicate these experiments by measuring capillary effects and surface tension in \cite{kernbach2023Pershin}, and the heat capacity in this work.

Since ortho/para- conversion is expected to have a weak character \cite{Miani04}, we use high-resolution active (with thermostat) and passive differential calorimeters. Active calorimeter is used for diathermic measurements with two thermal points and for providing a thermally stabilized environment for long-term measurements. Using passive differential calorimeter is argued by two reasons. First, it ensures better homogeneity of temperature distribution that is important for long-term measurements. Second, comparing active (thermal profile $21C \rightarrow 25C$) and passive (thermal profile $21C \rightarrow 17C$) calorimeters, we noted an increased level of fluctuations in thermostatic calorimeter, that can be caused by adding a thermal energy into the fluidic system. Since thermal fluctuations are observed not only after treatment, but also in their subsequent dynamics, we assume that the ongoing spin conversion possibly takes place in the long-term dynamics of fluidic systems.

\section{Setup and methodology}
\label{sec:setup}

\subsection{Considerations on heat capacity}
\label{sec:heat}

\begin{figure}
\centering
\subfigure{\includegraphics[width=0.49\textwidth]{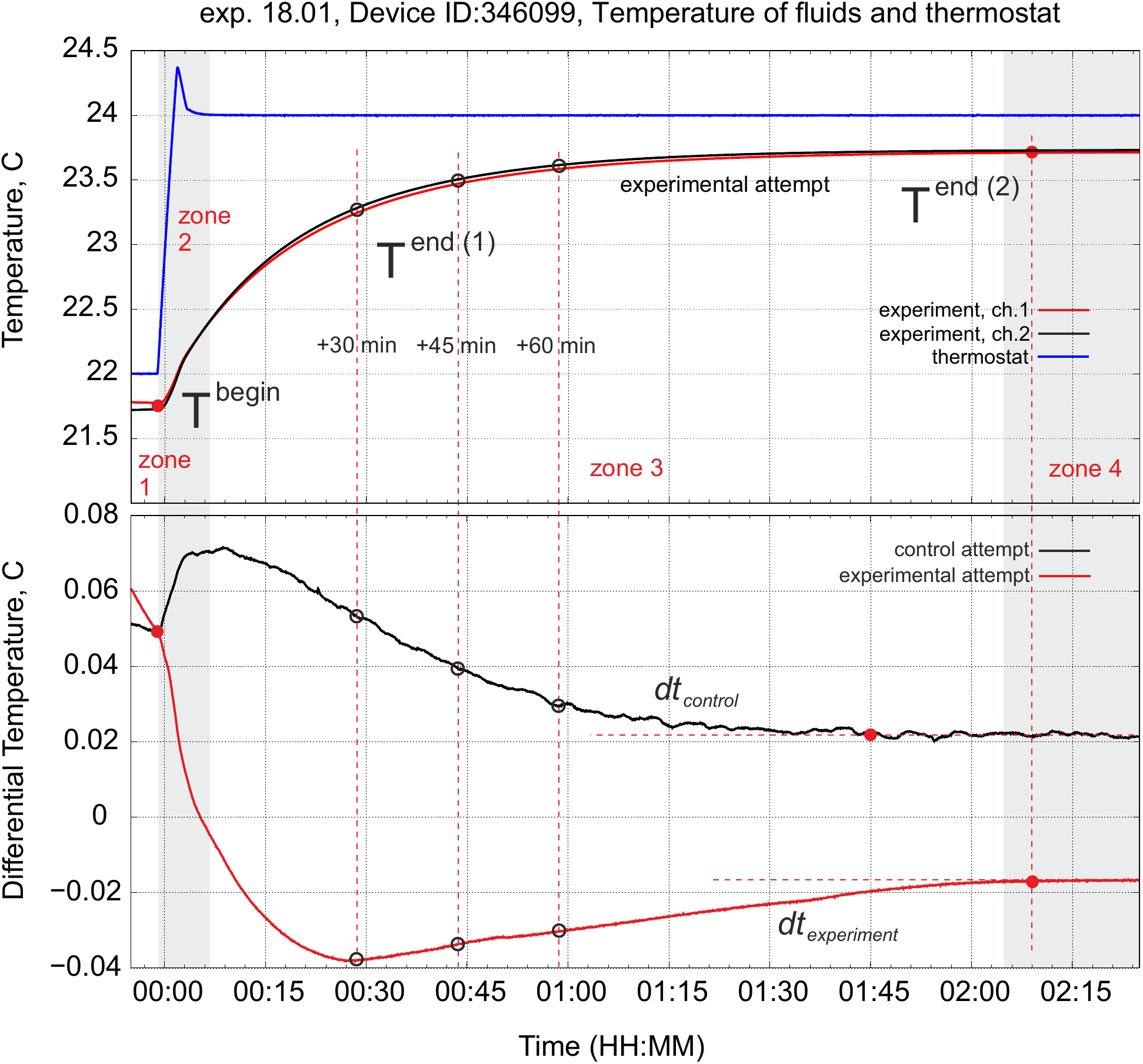}}
\caption{\textbf{(top)} Example of temperature dynamics in differential diathermic calorimeter with two settings of thermostat at 22$^\circ C$ and 24$^\circ C$, the experimental attempt is shown; \textbf{(bottom)} differential temperature dynamics of control and experimental attempts. Fluid samples are preheated in a water bath at 22$^\circ C$, their temperature difference at the point $T^{begin}$ is 0.0486$^\circ C$ and at $T^{end(2)}$ is 0.0232$^\circ C$ for control and -0.0169 $^\circ C$ for experimental attempt. $T^{end(1)}$ and  $T^{end(2)}$ are the type 1 or type 2 measurements using the schemes (\ref{eq:heat4}) or (\ref{eq:heat8}). The criteria for achieving the point $T^{end(2)}$ is a flat dynamics of differential temperature between samples in the zone 4. \label{fig:exampleCal}}
\end{figure}

The heat capacity and the change in temperature are related as 
\begin{equation}
\label{eq:heat}
Q=mC\Delta T,
\end{equation}
where $Q$ is a heat in joules (J), $m$ -- mass in kilograms (kg), $c$ -- specific heat capacity in J/kg$^\circ C$, $\Delta T$ -- temperature change in $^\circ C$. Using differential measurements with equal heat on left and right channels, we receive 
\begin{equation}
\label{eq:heat3}
\frac{T_L^{end}-T_L^{begin}}{T_R^{end}-T_R^{begin}}=k\frac{C_R}{C_L},
\end{equation}
where indexes $L$ and $R$ are two channels of differential calorimeter and $k=m_R/m_L$ with mass of water and containers. In ideal case $m_R=m_L$, in real measurements it introduces the inaccuracy related to filling the water into containers. Measurement of temperature in diathermic calorimeters $T_L^{end},T_L^{begin},T_R^{end},T_R^{begin}$ assumes that the transient dynamics for the first temperature point is finished, i.e. 
\begin{equation}
\label{eq:heat4}
T_L^{begin}=T_R^{begin},
\end{equation}
and $T_L^{end},T_R^{end}$ are measured after specific time (+30 min, +45 min, +60 min) in the transient dynamics of the second temperature point, which produces
\begin{equation}
\label{eq:heat5}
T_L^{end}=T_R^{end}+dt,
\end{equation}
where $dt$ is a differential temperature between channels, see Fig.~\ref{fig:exampleCal}.

Setting $\Delta T_R=\Delta T_{control}$, $\Delta T_L=\Delta T_{control}+dt$ and $C_R=C$, $C_L=C+dC$, where $dC$ a change of specific heat capacity in experimental sample, transforms (\ref{eq:heat3}) into    
\begin{equation}
\label{eq:heat6}
1+\frac{dt}{\Delta T_{control}}=k\frac{C}{C+dC}.
\end{equation}

Since we are mostly interested in the $\frac{dC}{C}$ value, finally we receive 
\begin{equation}
\label{eq:heat7}
\frac{dC}{C}= k \frac{1}{1+\frac{dt}{\Delta T_{control}}}-1.
\end{equation}

Typically, (\ref{eq:heat4}) is achieved after equalization in the water bath (prior to measurements) and in the calorimeter at $T^{begin}$. Essential variations between $T_L^{begin}$ and $T_R^{begin}$ caused by heating during excitation and by handling of samples, can affect the heat equality condition in (\ref{eq:heat3}) and lead to a significant inaccuracy of measurements. However, stabilizing the temperature around $T^{begin}$ for achieving (\ref{eq:heat4}) and then measuring $T^{end}$ for (\ref{eq:heat5}) requires considerable amount of time. Since the spin conversion in non-equilibrium state has own temporal dynamics and is expected to decrease with time, changes of heat capacity of isomers should be measured shortly after excitation.

The observation is that the condition (\ref{eq:heat5}) can be achieved from
\begin{equation}
\label{eq:heat8}
T_L^{begin} \approx T_R^{begin},
\end{equation}
considering that transient time is variable, it depends on a difference between $T_L^{begin}$ and $T_R^{begin}$, and measurements will be done in the steady-state area of $T^{end}$ after the second temperature point. This case is exemplified in Fig.~\ref{fig:exampleCal}. In other words, small variations between initial temperature of samples in the zone 1, caused e.g. by handling of samples, will be compensated in the zone 3; the criteria for achieving the steady-state zone 4 is a flat dynamics of differential temperature between samples. This consideration does not change (\ref{eq:heat7}) beside that $dt$ is measured after finishing the transient dynamics of $T^{end}$ instead of fixed time after $T^{begin}$.

Further, the schemes (\ref{eq:heat4}) or (\ref{eq:heat8}) are denoted as the type 1 or type 2 measurements with $T^{end(1)}$ and  $T^{end(2)}$, see Fig.~\ref{fig:exampleCal}. They can be considered as two alternative ways for measuring (\ref{eq:heat7}) -- with or without taking into account the transient dynamics in the zone 3 (see more in Discussion in Sec. \ref{sec:discussion}). Another reason for selecting the type 2 measurements is the initial difference between temperature of samples: for $dt>0.1$ $^\circ C$ it makes sense to use the scheme (\ref{eq:heat8}). 

\begin{figure*}
\centering
\subfigure[\label{fig:activethrmostat}]{\includegraphics[width=0.45\textwidth]{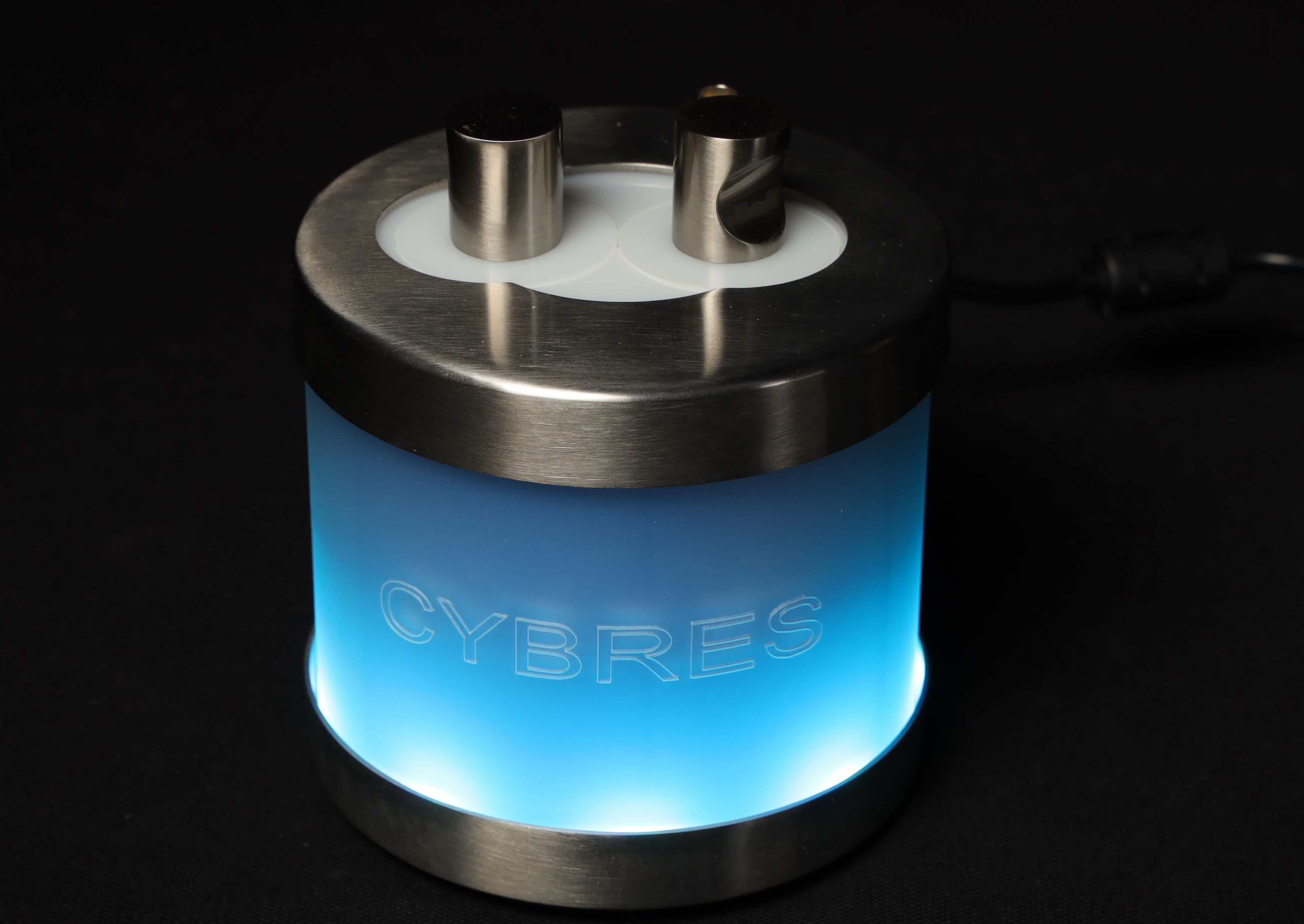}}~~~~~~
\subfigure[\label{fig:passiveCalorimeterA}]{\includegraphics[width=0.42\textwidth]{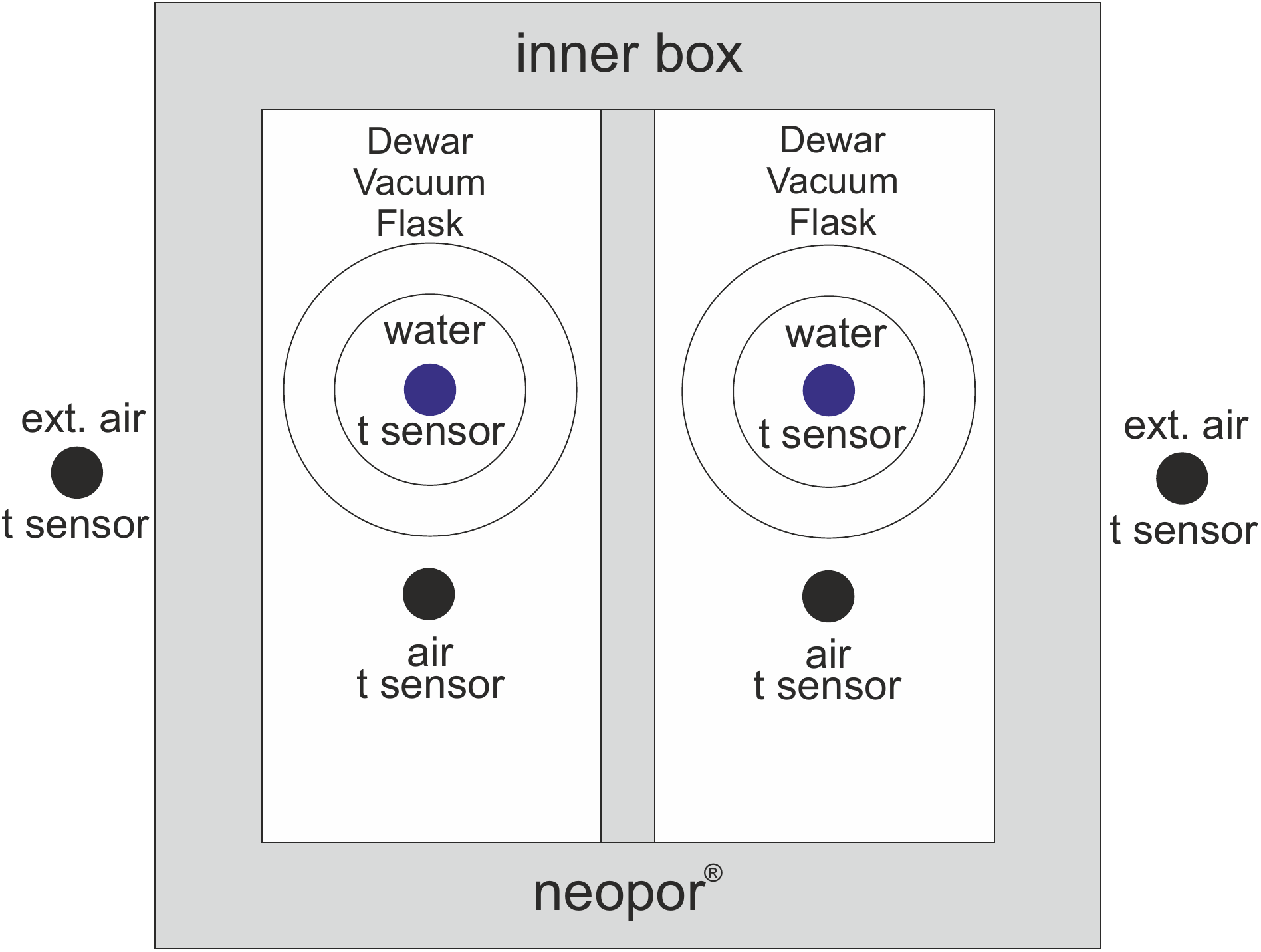}}
\subfigure[\label{fig:treatment}]{\includegraphics[width=0.55\textwidth]{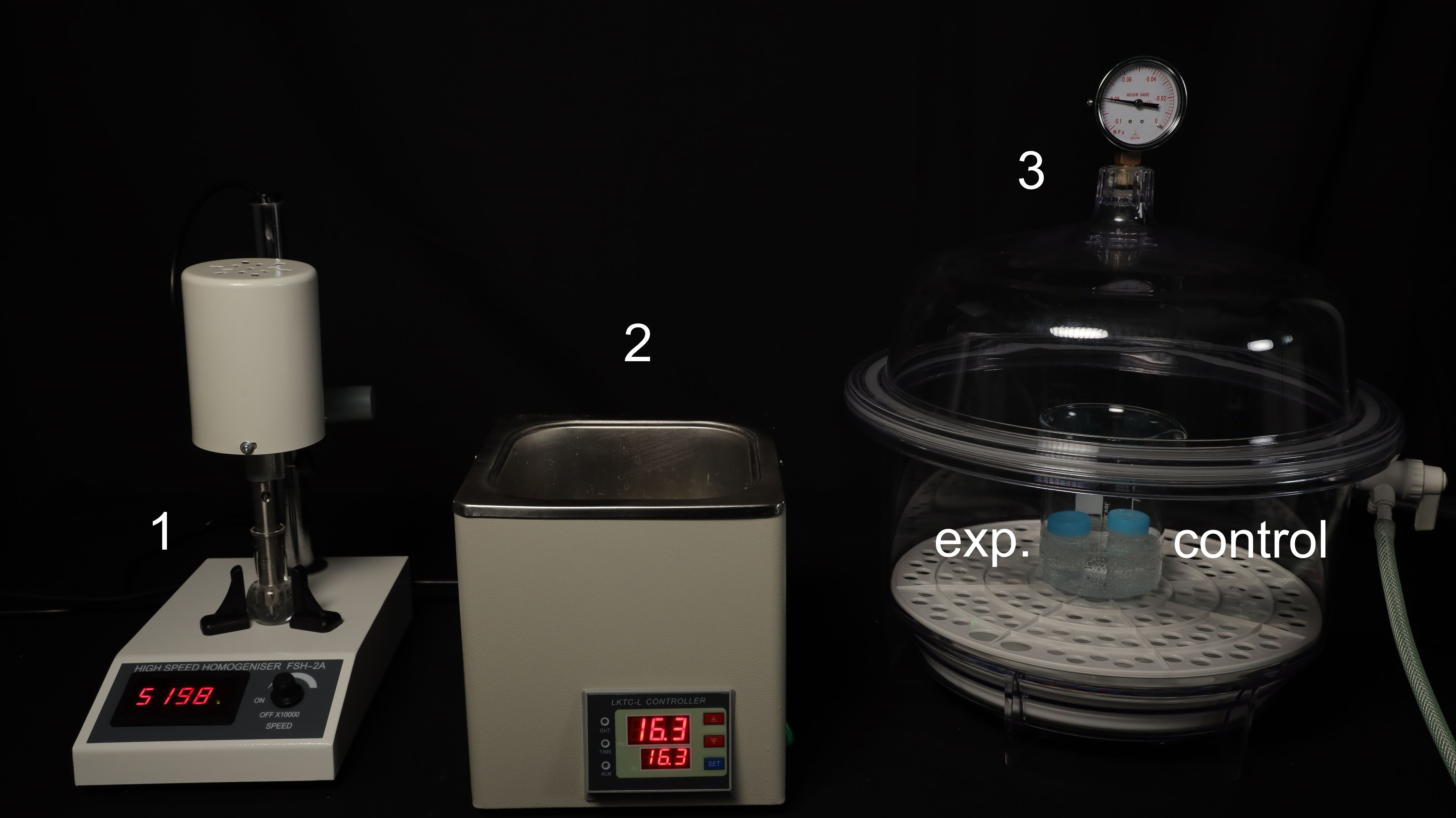}}~~
\subfigure[\label{fig:passiveCalorimeterB}]{\includegraphics[width=0.4\textwidth]{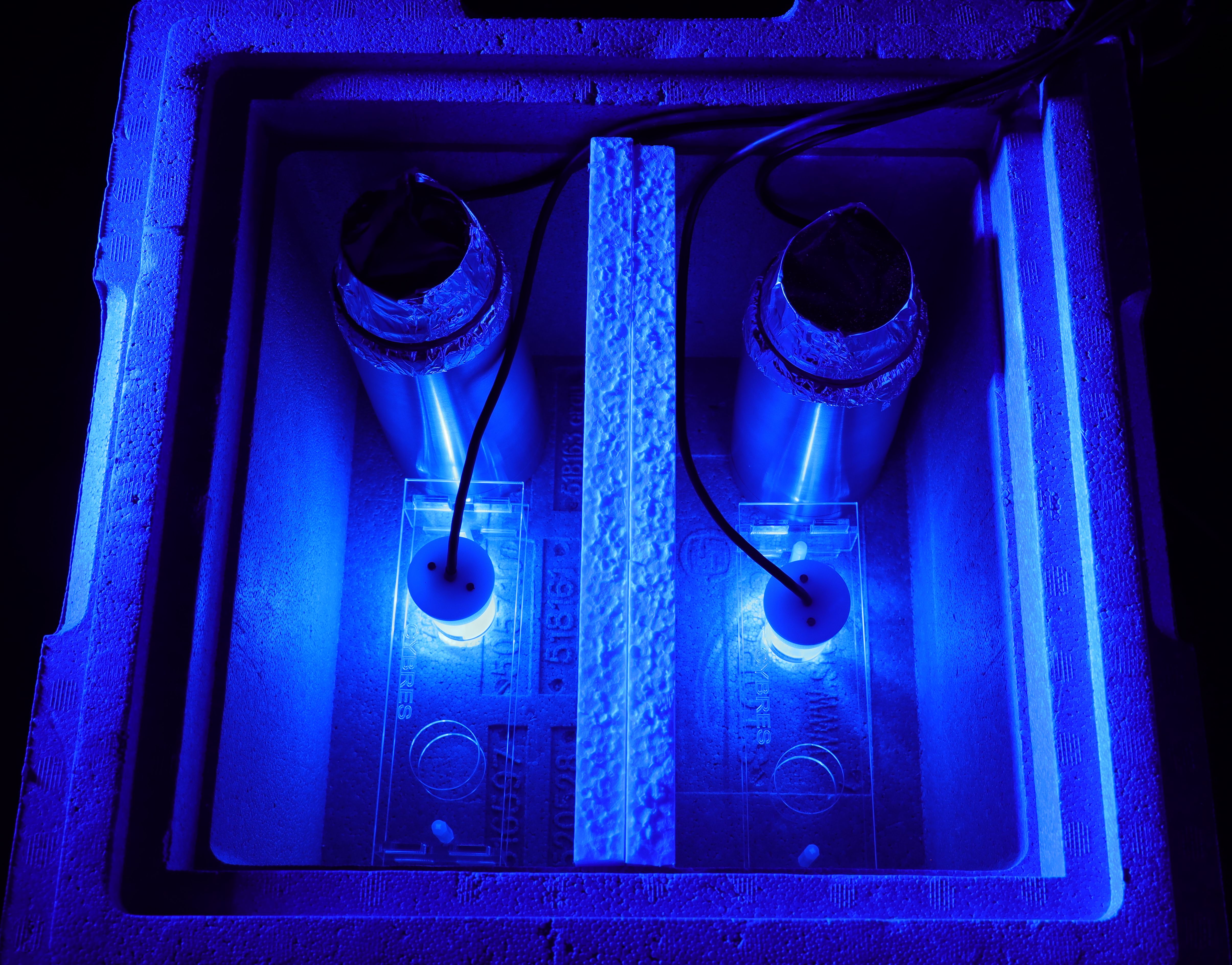}}
\caption{Experimental setup: \textbf{(a)} Differential calorimeter with active thermostat, the temperature feedback is obtained from the heating element, accuracy of thermal stabilization is about 0.002 $^\circ C$; \textbf{(b,d)} Passive differential calorimetric system with two dewar flasks with fluidic sensors and two air sensors in each section; \textbf{(c)} Setup for sample treatment: 1 -- hydrodynamic cavitation in mechanical way by high speed 10000-30000 rpm homogenizer (low speed rotation was used for weak mechanical excitation); 2 -- water bath for equalizing temperature of both samples and setting a specific initial measurement temperature; 3 -- degassing at -0.08 -- -0.09 MPa in small water bath. \label{fig:passiveCalorimeter}}
\end{figure*}

\textbf{Calibration of differential $t$ dynamics.} High resolution temperature dynamics is measured as time- and channel- differential in regard to initial $t$ points measured with the same sensors. From this point of view, there is no need to calibrate the $t$ sensors for accurate absolute temperature. However, even differential dynamics of water samples measured in different environmental and laboratory conditions demonstrate variations, see more in \cite{KernbachPh2022}. Therefore $dt$ in (\ref{eq:heat7}) should be calibrated to current control measurements. Each experiment should be conducted in two cases: with two unexcited fluids to calibrate $dt_{control}$ and with excited/unexcited fluids to measure $dt_{experiment}$. Finally, setting  
\begin{equation}
\label{eq:heat9}
dt=dt_{experiment}-dt_{control}
\end{equation}
in (\ref{eq:heat7}) will deliver the required $\frac{dC}{C}$. Note, that the transformation (\ref{eq:heat9}) is considered as calibration of the zero level from control attempts, whereas the $\Delta T_{control}$ in (\ref{eq:heat7}) is related to a control channel (unexcited sample) in experimental attempts.

\textbf{Sources of inaccuracy in differential calorimetric measurements.} Typical calorimetric inaccuracies are described e.g. in \cite{Calorimetry14}, from them we see four main sources: 1) errors in measuring temperature, 2) inaccurate filling of fluids in measurement containers, 3) inhomogeneous distribution of temperature in the calorimeters and finally, 4) variation of $dt$ in the zone 4, see Fig.~\ref{fig:exampleCal}. Since $dt$ is channels-differential and $\Delta T_{control}$ is time-differential temperature, both relative values can be measured with a high resolution and accuracy below $\pm 10^{-5}$ $^\circ C$. Considering density of water as 1 g/ml at 25 $^\circ C$, inaccuracy of filling in 0.05ml produces the variation in $k$ as $\pm 0.05$, i.e. $\frac{dC}{C}=0.(01) \pm 0.0005$ or 0.05\%. Typical variation of $dt$ in the zone 4 is $<0.001$, which also produces $\frac{dC}{C}=\pm 0.0005$ or 0.05\%. Thus, the largest worst-case inaccuracy of $\pm0.1\%$ is adopted for further considerations. Non-uniform temperature distribution affects transient dynamics and is a typical problem for active thermostats due to uncontrollable factors related to thermal insulation, conduction of heat to samples, etc. To resolve this issue, we compare transient and steady-state dynamics and perform several experiments with different temperature profiles.

\subsection{Setup}

\begin{table*}[h]
\begin{center}
\caption{\small Table of results as mean$\pm$StDev, the expression (\ref{eq:heat10}) is used for calculating $dt$, $N$ is a number of attempts. \label{tab:results} \vspace{-3mm}}
\fontsize {9} {10} \selectfont
\begin{tabular}{
p{0.5cm}@{\extracolsep{3mm}}|
p{1.2cm}@{\extracolsep{3mm}}
p{1.2cm}@{\extracolsep{3mm}}
p{1.2cm}@{\extracolsep{3mm}}
p{1.2cm}@{\extracolsep{3mm}}
p{1.2cm}@{\extracolsep{3mm}}
p{1.2cm}@{\extracolsep{3mm}}
p{1.2cm}@{\extracolsep{3mm}}
p{1.2cm}@{\extracolsep{3mm}}
p{1.2cm}@{\extracolsep{3mm}}
p{1.2cm}@{\extracolsep{3mm}}
p{1.2cm}@{\extracolsep{3mm}}
p{1.2cm}@{\extracolsep{3mm}}
}\hline \hline
    & \multicolumn{9}{c|}{Type 1} & \multicolumn{3}{c}{Type 2} \\\hline  
    & \multicolumn{3}{c|}{30 min} & \multicolumn{3}{c|}{45 min}  & \multicolumn{3}{c|}{60 min} &  \multicolumn{3}{c}{} \\\hline 
N    & $\Delta T$ $^\circ C$ & $dt$ $^\circ C$    & $\frac{dC}{C}$ & $\Delta T$ $^\circ C$ & $dt$ $^\circ C$ & $\frac{dC}{C}$ & $\Delta T$ $^\circ C$ & $dt$ $^\circ C$ & $\frac{dC}{C}$ & $\Delta T$ $^\circ C$ & $dt$ $^\circ C$ & $\frac{dC}{C}$ \\\hline 
 20 
& 1.5549 $\pm$0.0466 & -0.0821 $\pm$0.0564 & 0.0572 $\pm$0.0418
& 1.7690 $\pm$0.0525 & -0.0696 $\pm$0.0458 & 0.0417 $\pm$0.0288  
& 1.8770 $\pm$0.0539 & -0.0608 $\pm$0.0359 & 0.0392 $\pm$0.0206 
& --- & --- & ---\\[3mm]
 24  &---&---&---&---&---&---&---&---&---
& 1.9865 $\pm$0.0683 & -0.0403 $\pm$0.0152 & 0.0208 $\pm$0.0082
\\
\hline \hline 
\end{tabular}
\end{center}
\end{table*}

\textbf{Calorimetric measurements} have been conducted with two different systems: with active thermostat, see Fig. \ref{fig:activethrmostat} and fully passive, see Figs. \ref{fig:passiveCalorimeterA}, \ref{fig:passiveCalorimeterB}. Both systems are differential and have air temperature sensors to determine environmental conditions. Fluid sensors are TNC thermistors immersed into the fluids. Air temperature is measured by the same thermistors and by precision semiconductor (LM35) sensors. Power supply is monitored by independent sensors, PCB with electronic component is thermostabilized by active thermostat. With 24 bit ADC, the \textit{t} resolution is $<10^{-5}$ $^\circ C$ (relative temperature). Active thermostat uses the temperature feedback from heating elements (not from water samples), this provides a stable environmental temperature for samples and does not distort their own thermal dynamics. The upper part of active calorimeter is additionally thermally insulated; the whole system is placed into neopor container like shown in Fig. \ref{fig:passiveCalorimeterB}. The thermostat and \textit{t} measurement are two different, fully uncoupled, systems with independent power supply; accuracy of thermostat is about 0.002 $^\circ C$. 

\textbf{Preparation of laboratory.} Active and passive calorimeters are installed in specially prepared laboratory. Windows and doors are covered by polymer firms, all heat producing elements (e.g. PCs) are removed, all heaters are turned off. Measurements are conducted remotely without direct human intervention. Under these conditions, the circadian rhythm was typically around 0.3 $^\circ C$ for 96 hours, see Fig. \ref{fig:thermalVariations}, which is used for the passive calorimeter.

\textbf{Preparation of samples.} Experimental samples are treated by hydrodynamic cavitation produced in mechanical way by a standard laboratory homogeniser (FSH-2A) with 10000 rpm, see Fig. \ref{fig:treatment}. Similarly to \cite{kernbach2023Pershin}, several attempts have been conducted with the ultrasound 40kHz (ultrasonic cleaners) and 1.7MHz (ultrasonic vaporizer) excitation. However, this excitation essentially heats the samples (over 40-45 $^\circ C$), thus such attempts have been canceled. Since mechanical excitation also heats fluidic samples, the treatment time (typically 5-10 min) was selected so that to keep the temperature below $25$ $^\circ C$ (initial temperature about 17 $^\circ C$). After that both samples have been degassed at -0.09 MPa in glass containers with large surface area (height of fluid is about 2 mm) without stirring to avoid excitation of control samples. Finally, the temperature of control and experimental samples equalized in water bath and brought to a specified initial measurement temperature, typically 20-22$^\circ C$. The overall treatment time after excitation was kept to about 10-15 min. Water samples are filled into 15 ml containers with laboratory dropper pipette providing accuracy of about 0.05 ml. In parallel, the same samples are filled in 10 ml contained for testing capillary effects \cite{kernbach2023Pershin}. Double distilled water with initial conductivity $<0.05 \mu S/cm$ was used in experiments.

\section{Results}

\begin{figure}[ht]
\centering
\subfigure{\includegraphics[width=0.49\textwidth]{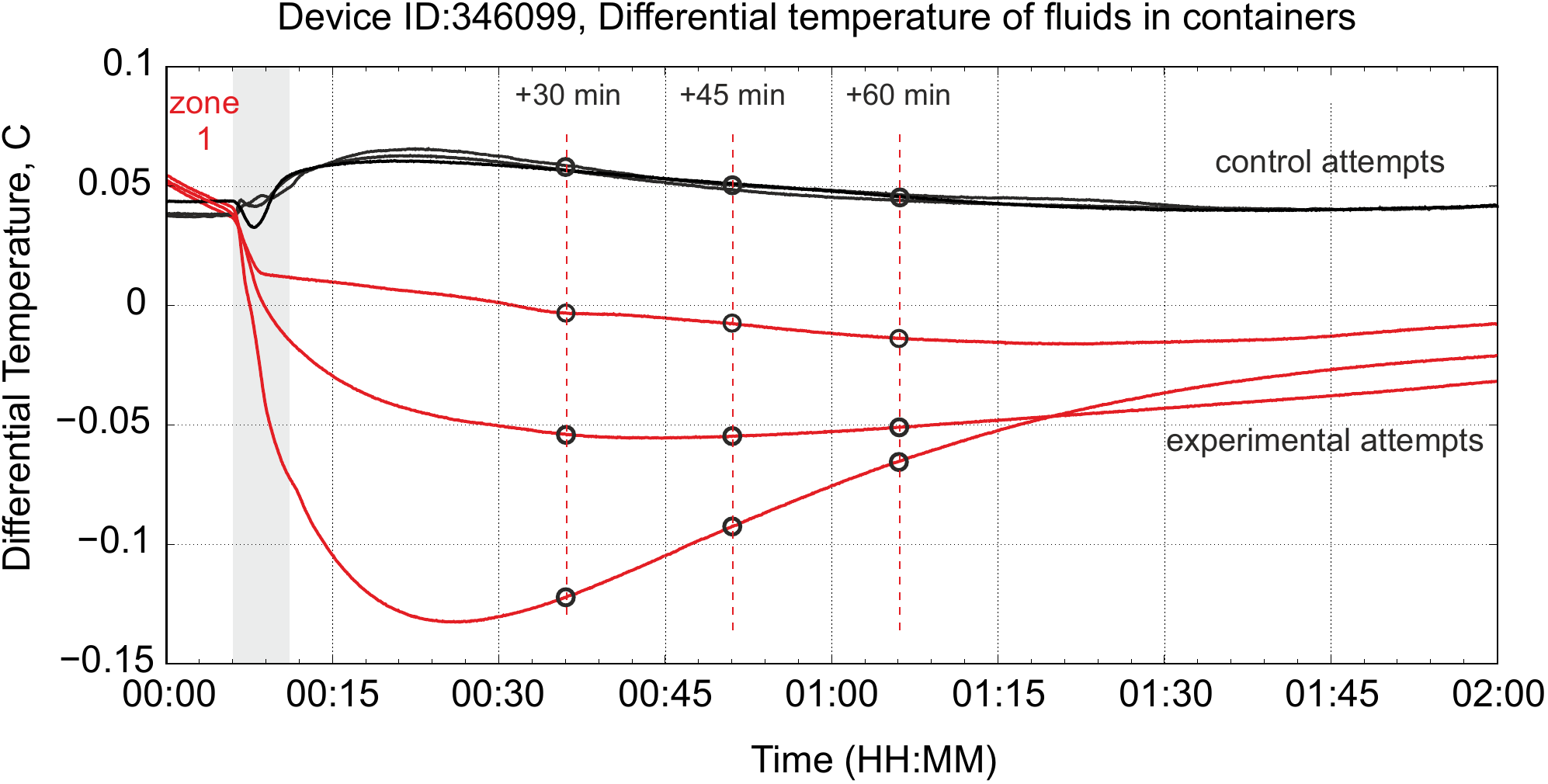}}
\caption{Comparison of several control (black) and experimental (red) attempts with type 1 measurements; control attempts demonstrate less variations between measurements. Time points 30 min, 45 min, and 60 min after setting of the second temperature point are shown by circles. \label{fig:comparison}}
\end{figure}

\textbf{1. Control measurements} are conducted immediately before or after each experimental attempt to calculate (\ref{eq:heat9}). Fig. \ref{fig:comparison} compares several control and experimental measurements, we observe less variations between measurements in control attempts. Low variability of control attempts enables to rewrite (\ref{eq:heat9}) as
\begin{equation}
\label{eq:heat10}
dt_i=dt_{i~experiment}-\bar{dt}_{control}
\end{equation}
where $\bar{dt}_{control}$ is a mean of all control attempts at the corresponding time interval. The expression (\ref{eq:heat10}) allow reducing variability of $\frac{dC}{C}$ value in (\ref{eq:heat7}).     

\begin{figure}
\centering
\subfigure[\label{fig:shortTermA}]{\includegraphics[width=0.49\textwidth]{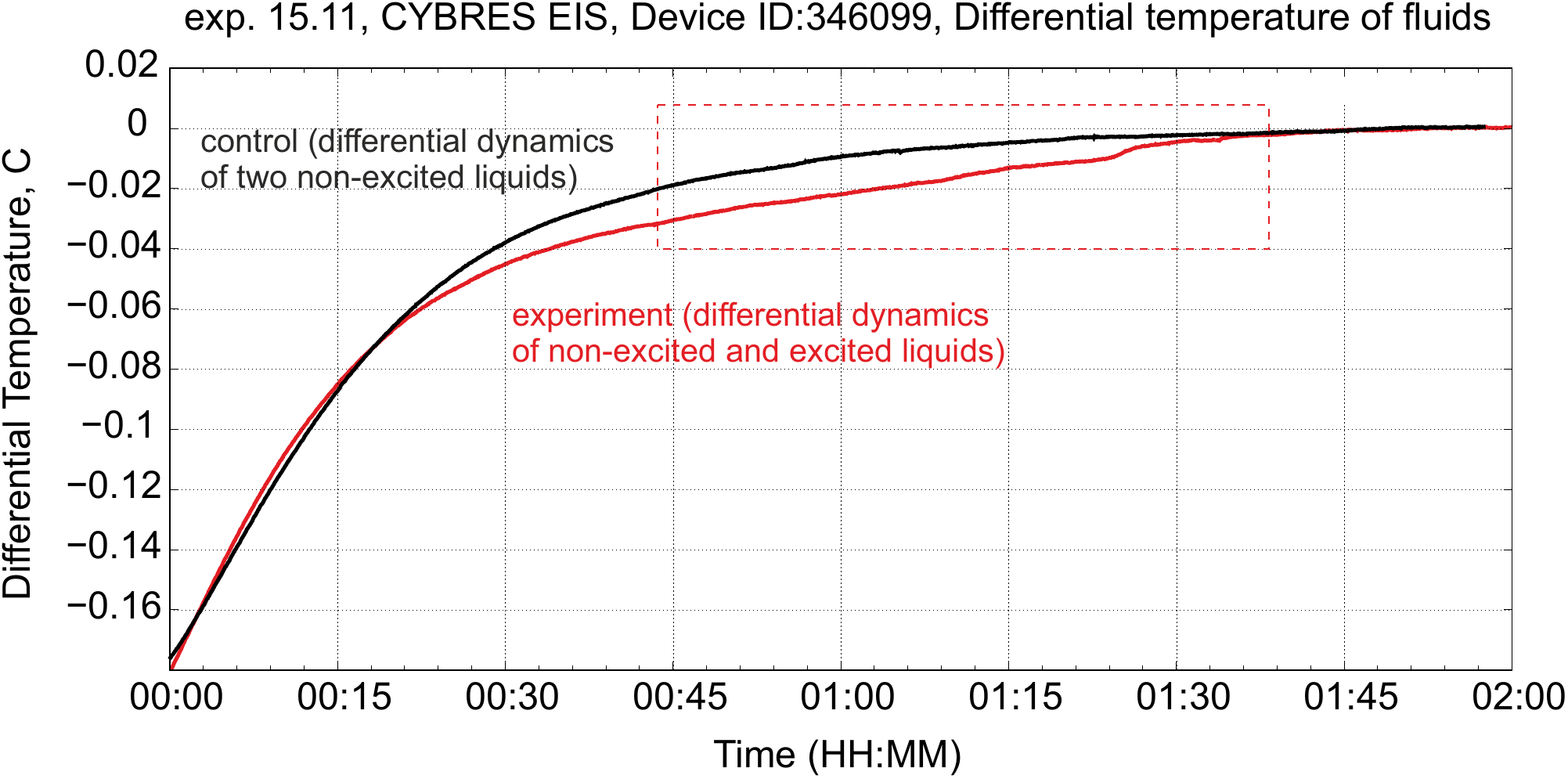}}
\subfigure[\label{fig:shortTermB}]{\includegraphics[width=0.49\textwidth]{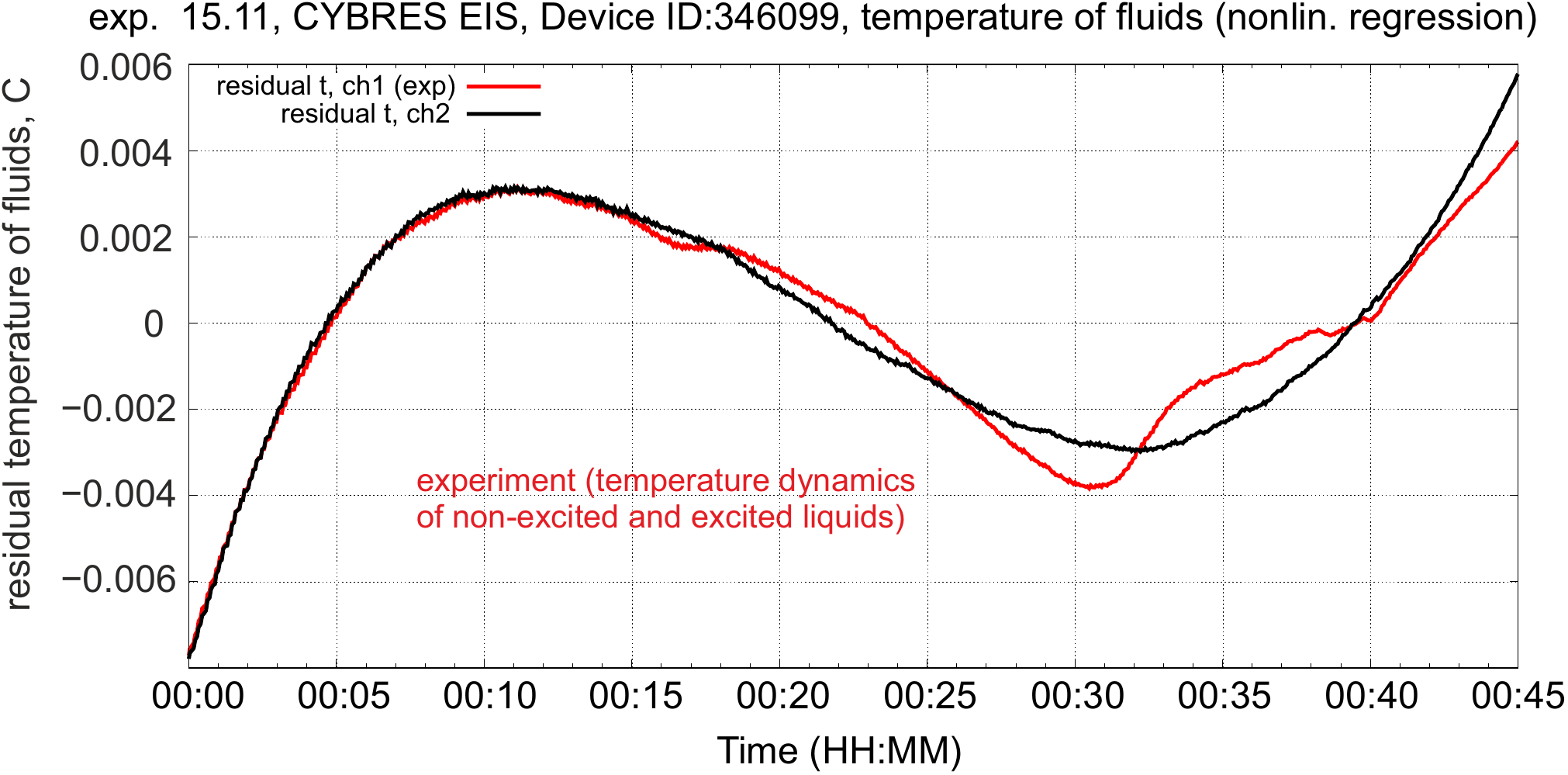}}
\caption{Example of short-term fluctuations after treatment (21$^\circ C$ to 25$^\circ C$ profile): \textbf{(a)} Differential temperature dynamics of control (two equal samples without treatment) and experimental (one sample is excited) attempts, shown is the transient phase before achieving the stable temperature; \textbf{(b)} Removing trend by nonlinear regression from experimental attempt -- residual temperature dynamics of excited and unexcited liquids. \label{fig:shortTerm1}}
\end{figure}

\textbf{2. Type 1 and type 2 measurements.} All experimental attempts are conducted with the type 1 and type 2 measurements. If the temperature equalization and the handling produce initial $dt>0.1$$^\circ C$, only the steady-state dynamics is measured. Typically, experimental attempts demonstrate different transient dynamics around the first temperature point than control attempts with two unexcited samples, see Figs. \ref{fig:exampleCal} and \ref{fig:comparison} in the zone 1. The same effect is also shown in Fig. \ref{fig:shortTermA}. After the second temperature point, all experimental measurements demonstrate higher absolute values compared to the control, however they vary between attempts. Table \ref{tab:results} collects several results of type 1 and type 2 measurements, typically we compare several temperature points -- 30 min, 45 min, and 60 min after setting of the second temperature point in the thermostat. Considering about 20 min of initial handling and equalizing temperature in the zone 1, it produces about 50 min, 66 min, and 80 min after excitation. 

\begin{figure}
\centering
\subfigure[]{\includegraphics[width=0.49\textwidth]{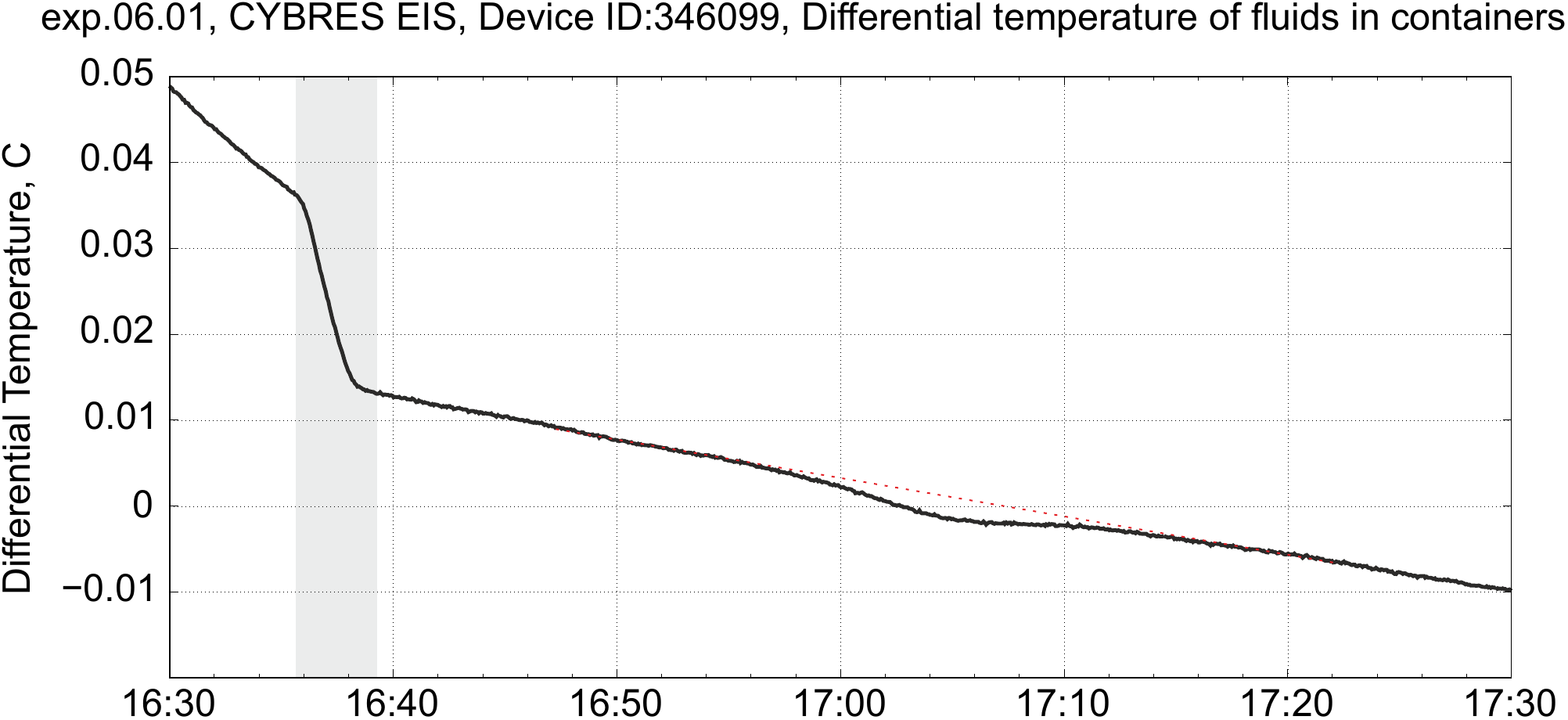}}
\subfigure[]{\includegraphics[width=0.49\textwidth]{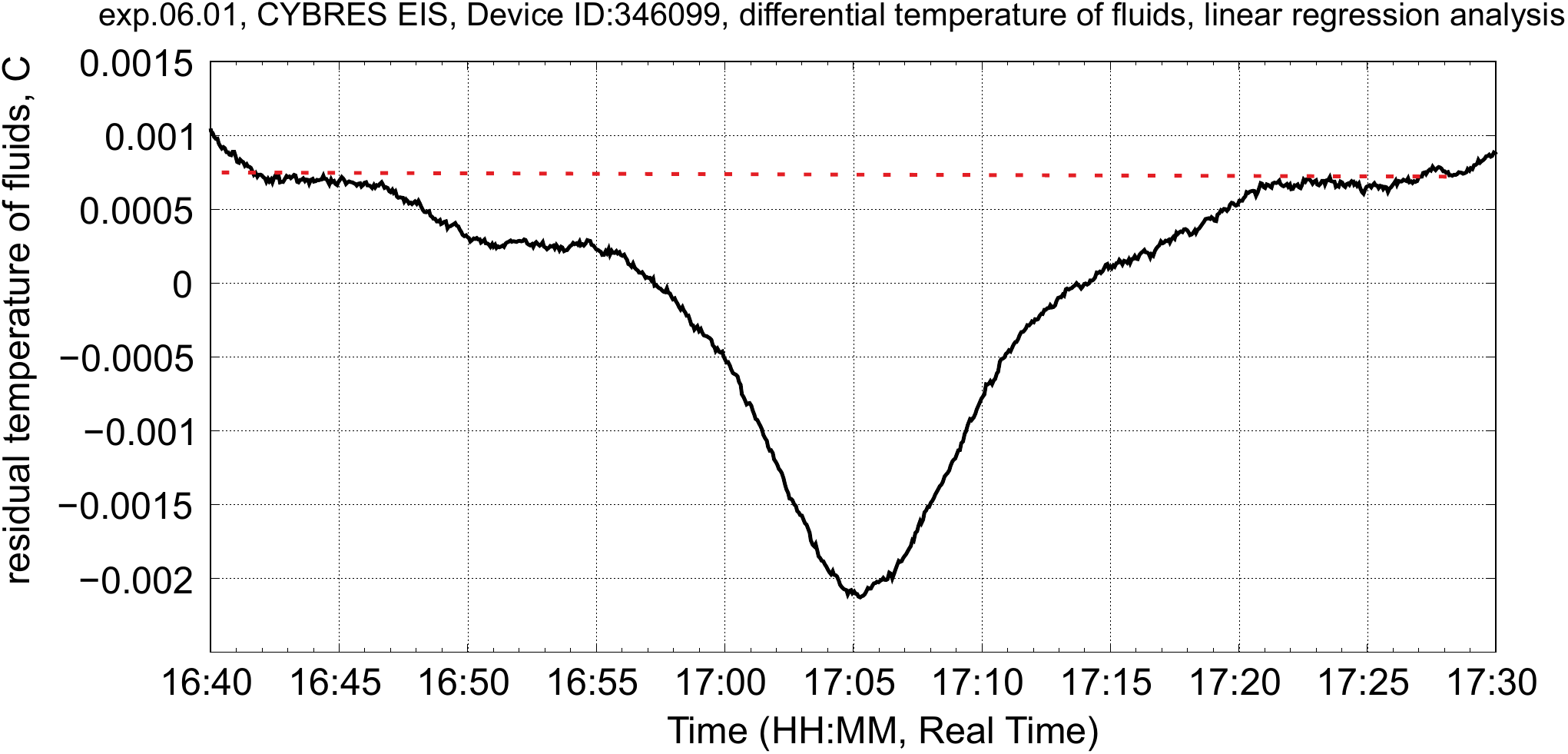}}
\caption{Example of short-term fluctuations after treatment (22$^\circ C$ to 24$^\circ C$ profile). \textbf{(a)} Differential dynamics of experimental attempt; \textbf{(b)} Linear regression of the region after setting the second temperature point for removing the trend, the temperature fluctuation of about 0.02$^\circ C$ for 20 min is shown. \label{fig:shortTerm2}}
\end{figure}

\textbf{3. Short-term thermal fluctuations} after treatments were observed in active and passive calorimeters at the level of 0.001-0.002$^\circ C$ for 20-40 min in excited liquid. Typically they occur in the transient dynamics when reaching the first temperature point, see Fig. \ref{fig:shortTermB}, or by following the diathermic profile after the second temperature point, see Fig. \ref{fig:shortTerm2}. We did not notice any particular regularity or special dependencies in the occurrence of these fluctuations.

\begin{figure}
\centering
\subfigure[\label{fig:longTermReactionA}]{\includegraphics[width=0.49\textwidth]{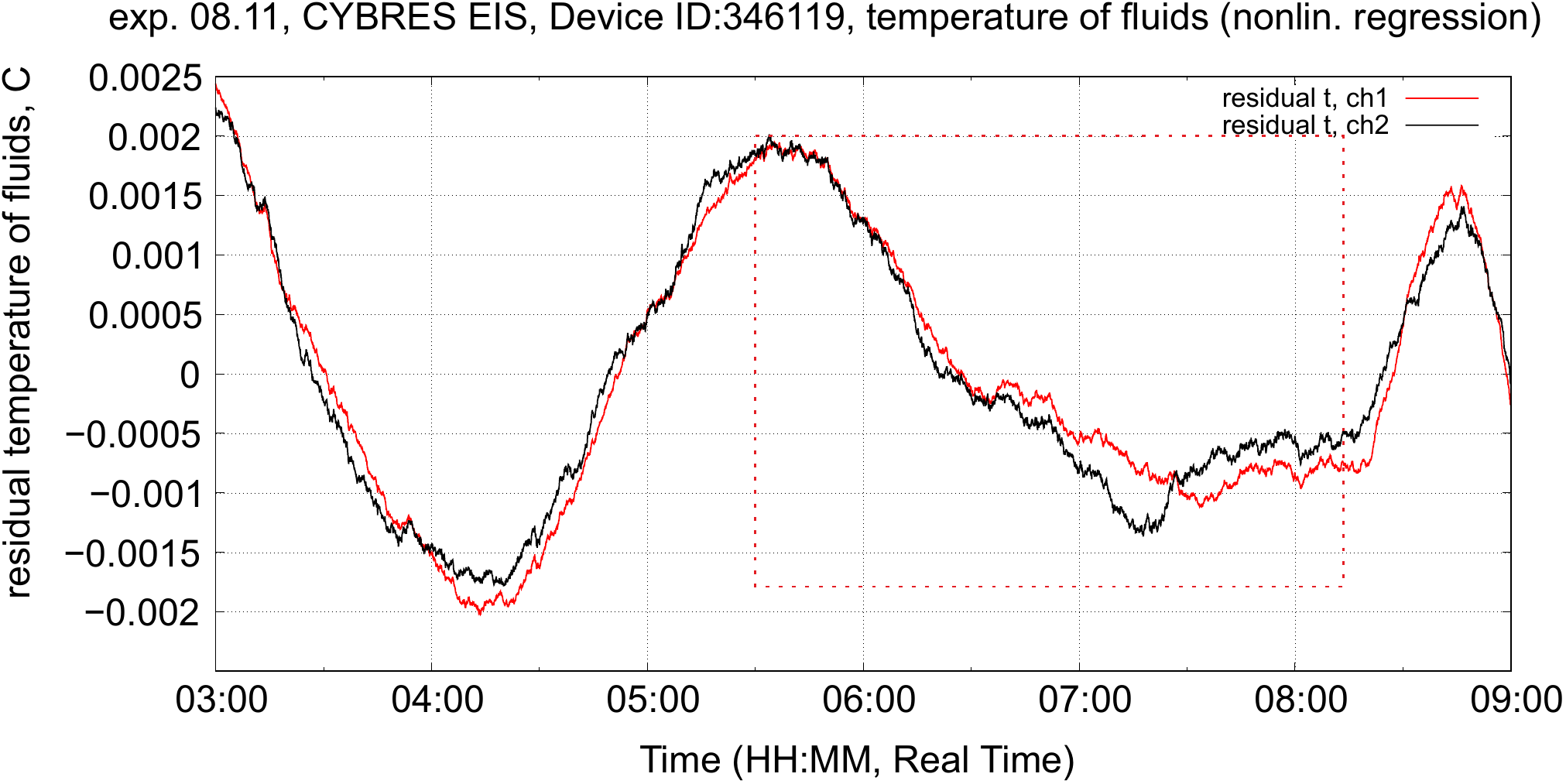}}
\subfigure[\label{fig:longTermReactionB}]{\includegraphics[width=0.49\textwidth]{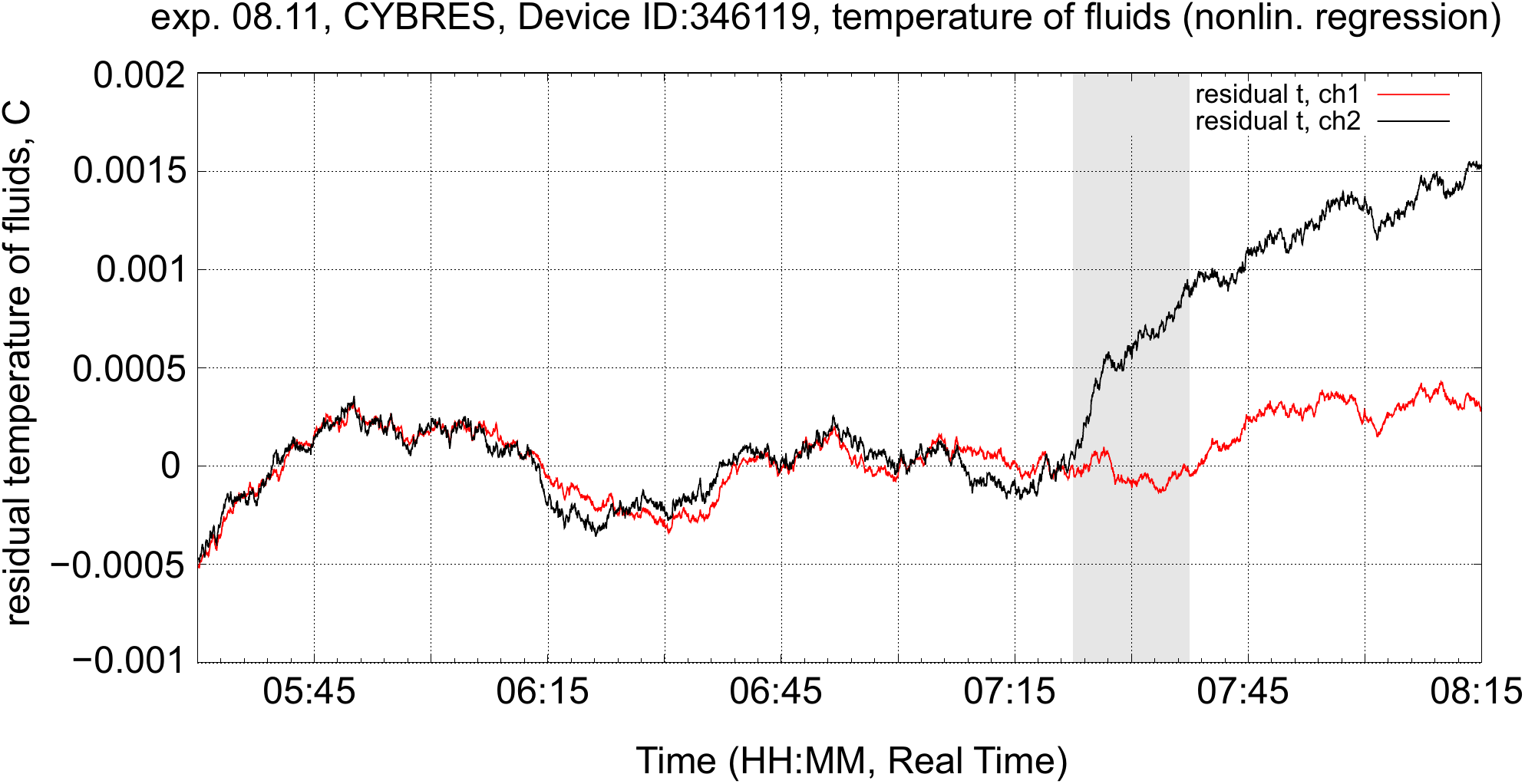}}
\subfigure[\label{fig:longTermReactionC}]{\includegraphics[width=0.49\textwidth]{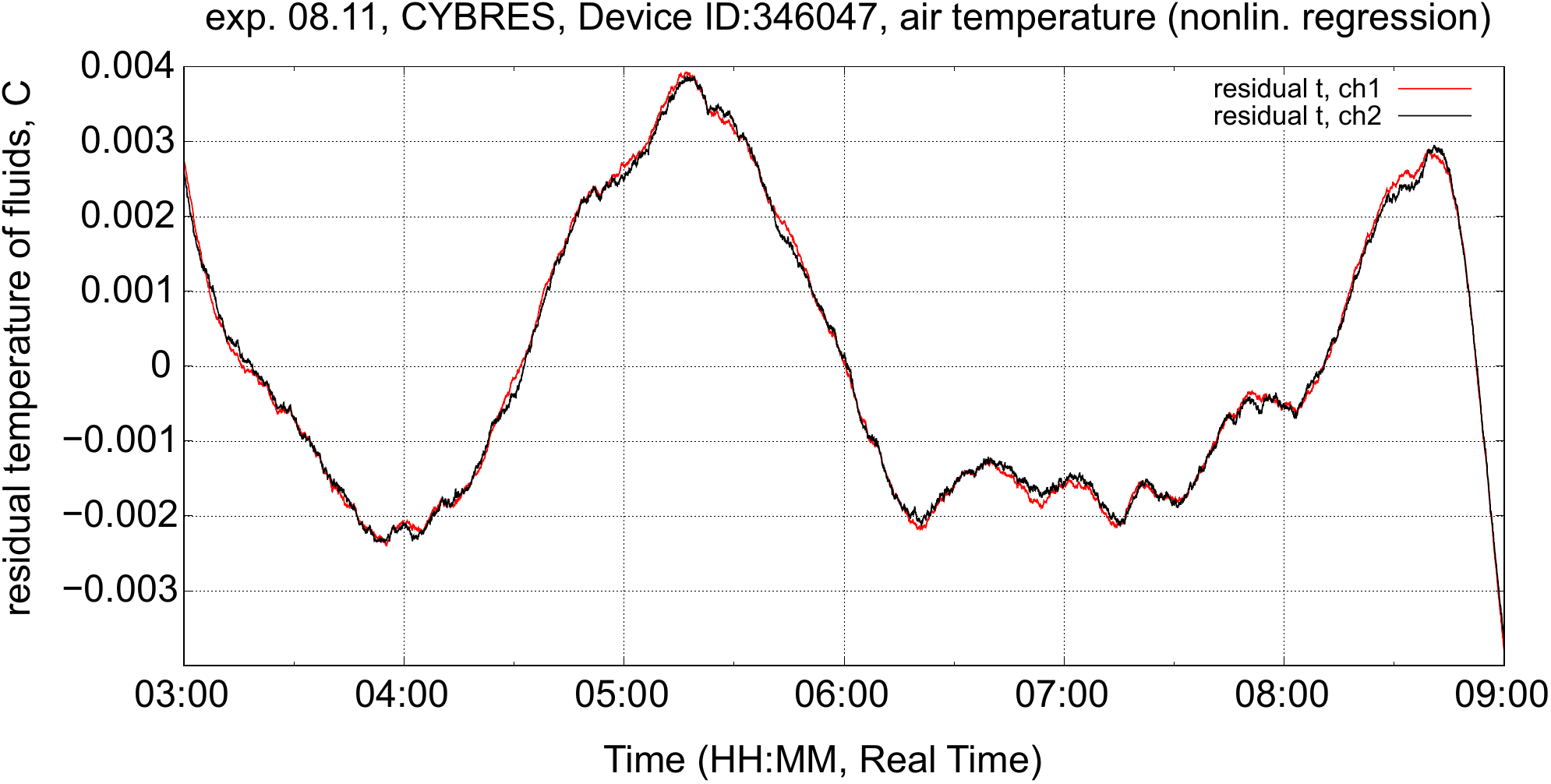}}
\caption{Example of long-term fluctuations, the passive calorimeter: \textbf{(a)} temperature of samples during six hours period, two days after the excitation by hydrodynamic cavitation,  temperature trend is removed by non-linear regression; \textbf{(b)} the selected region of temperature from (a); \textbf{(c)} Temperature dynamics of air sensors, no fluctuations between channels are measured. \label{fig:longTermReaction}}
\end{figure}

\begin{figure}
\centering
\subfigure[]{\includegraphics[width=0.49\textwidth]{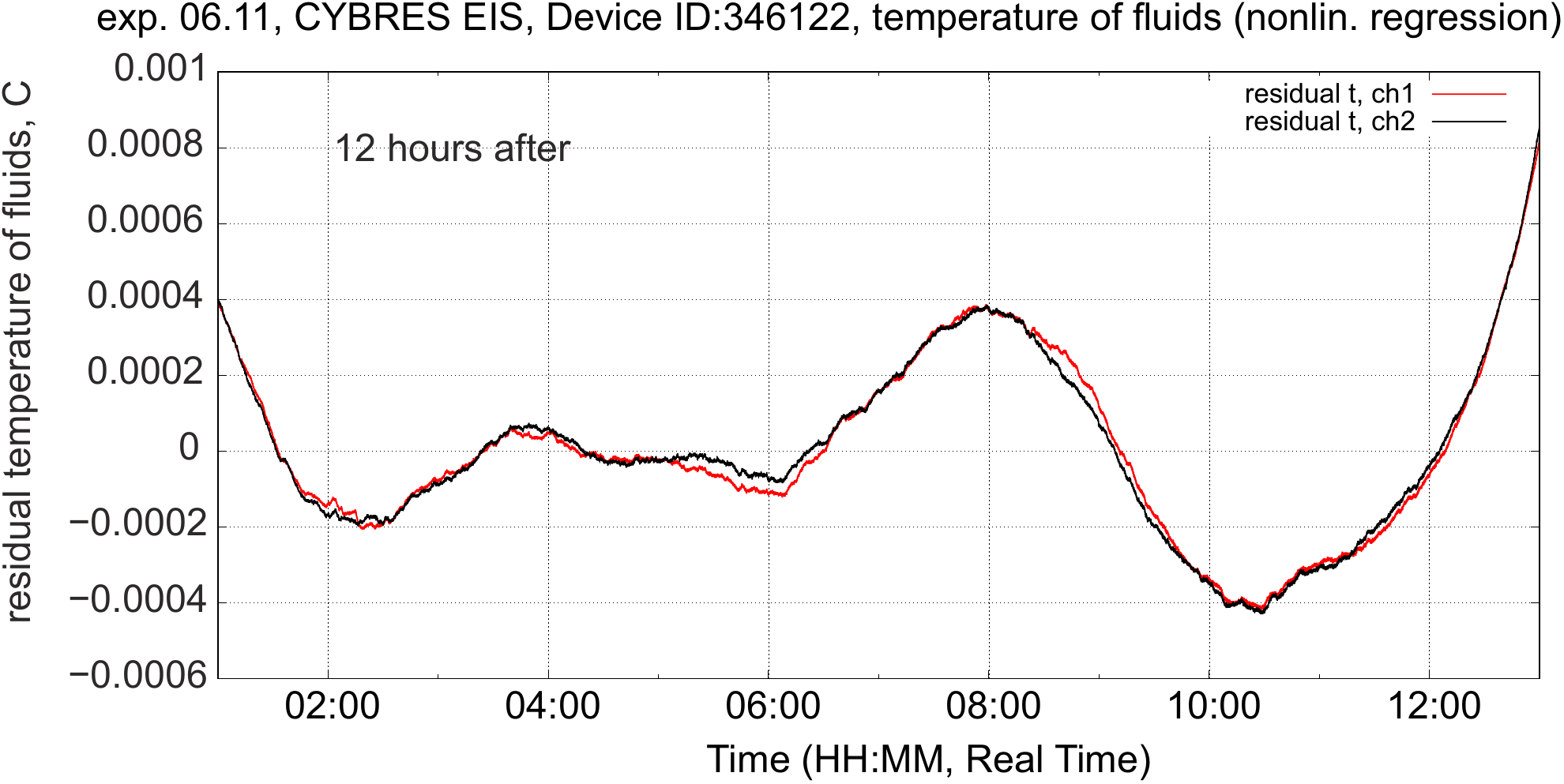}}
\subfigure[]{\includegraphics[width=0.49\textwidth]{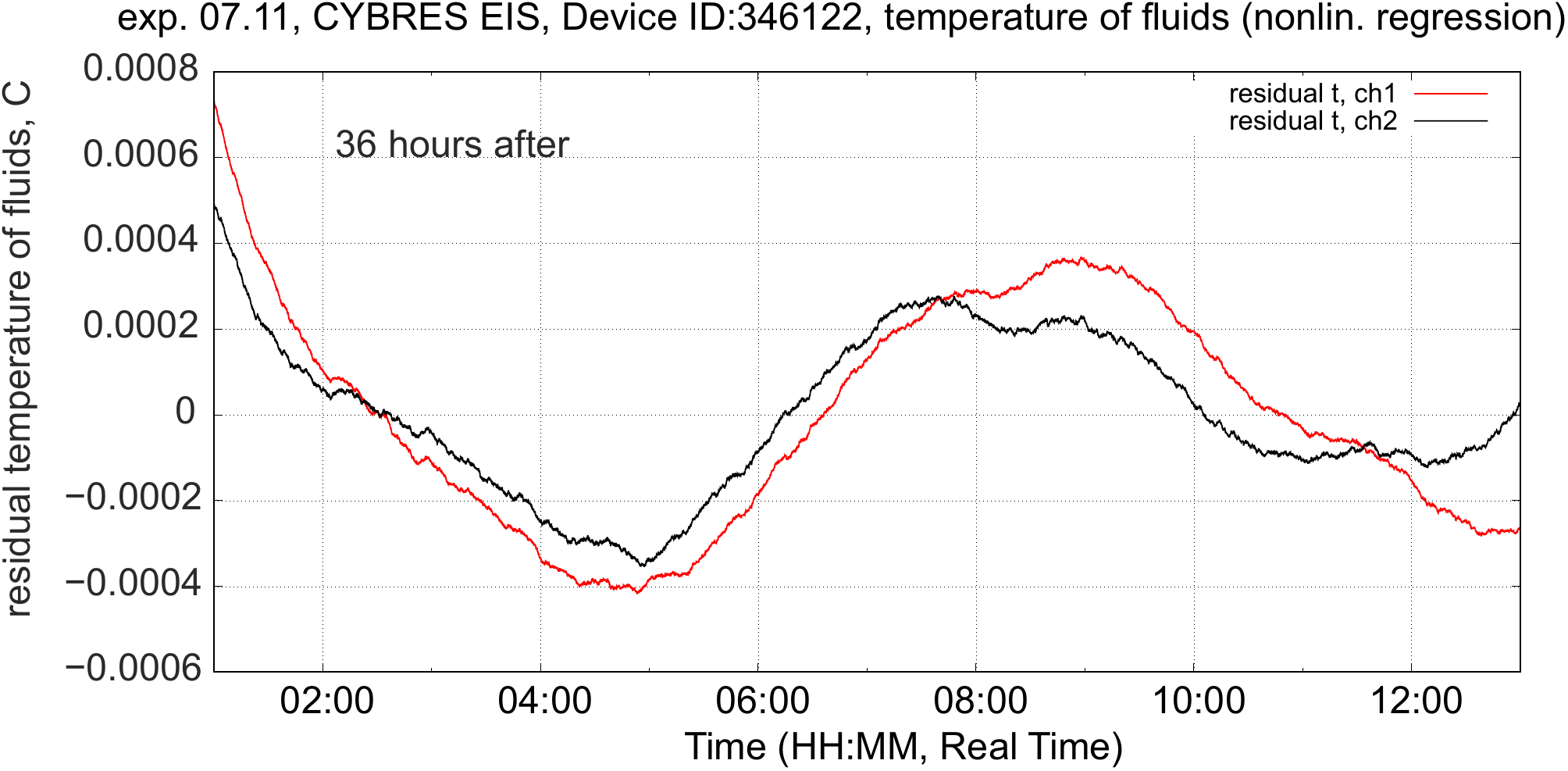}}
\subfigure[]{\includegraphics[width=0.49\textwidth]{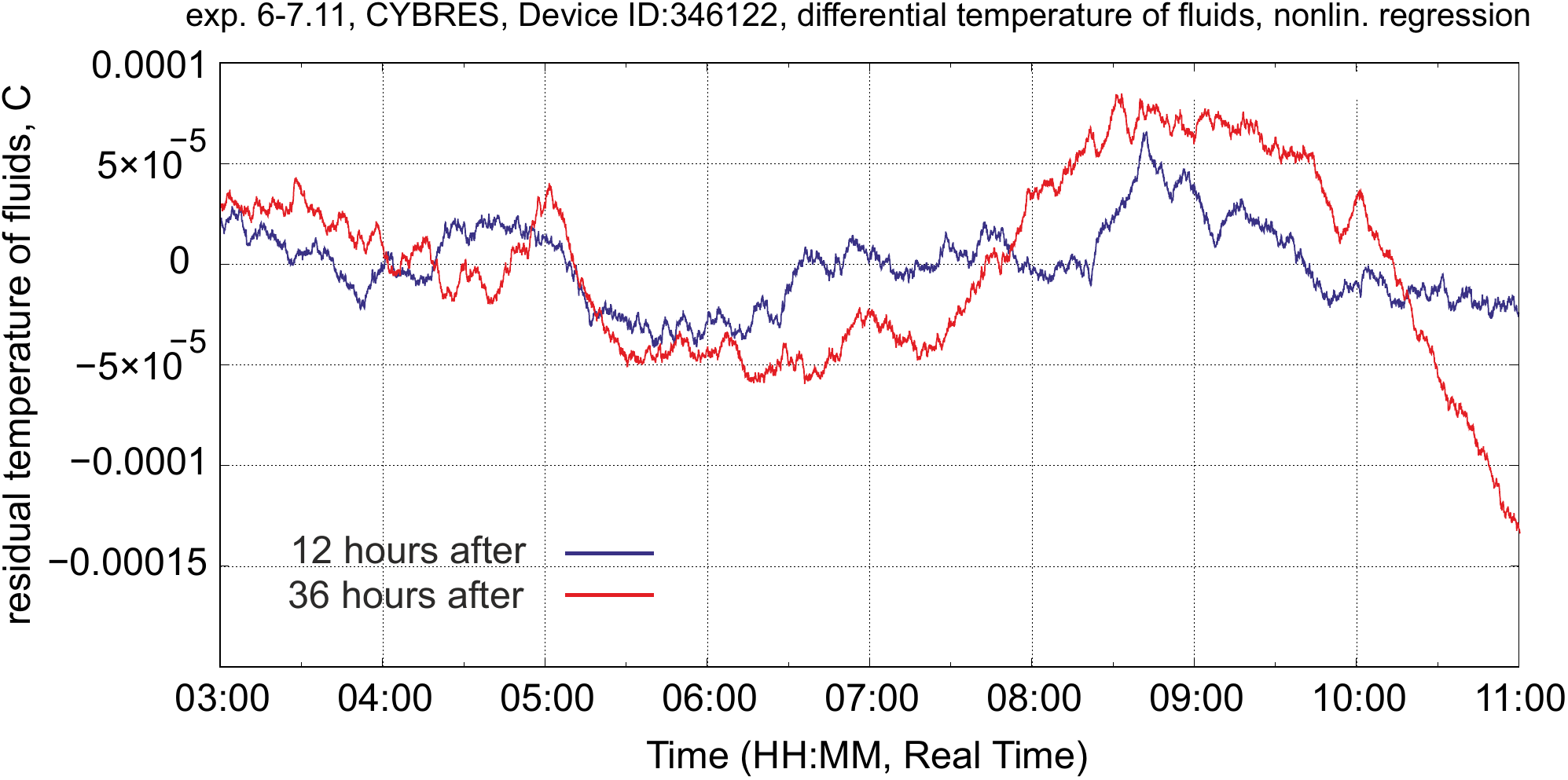}}
\caption{Long-term temperature dynamics, the passive calorimeter, trend is removed by regression: \textbf{(a)} 12 hours after the excitation by hydrodynamic cavitation; \textbf{(b)} the same samples 36 hours after the excitation; \textbf{(c)} Comparison of differential temperature in these two cases. \label{fig:transient_time}}
\end{figure}

\textbf{4. Long-term fluctuations} were measured in active and passive calorimeters on intervals up to 10 days after excitation of experimental sample. Two main effects have been discovered. 
First, thermal fluctuations observed in initial phase after treatments continue also in a long-term dynamics but with smaller amplitude (typically at the level of $10^{-3}$--$10^{-4}$ $^\circ C$ of relative temperature). Figures \ref{fig:longTermReactionA}, \ref{fig:longTermReactionB} provide an example of two days after excitation, where we observe a symmetry breaking between fluidic channels, but no differences between air temperature sensors (i.e. such fluctuation is caused by internal reasons in aqueous solutions), see Fig. \ref{fig:longTermReactionC}. Measurements demonstrate that the level of fluctuation is increasing over time. For instance, comparing the temperature dynamics 12 hours after excitation and 36 hours after excitation, see Fig. \ref{fig:transient_time}, larger fluctuations are observed in the second case.    

\begin{figure}
\centering
\subfigure{\includegraphics[width=0.49\textwidth]{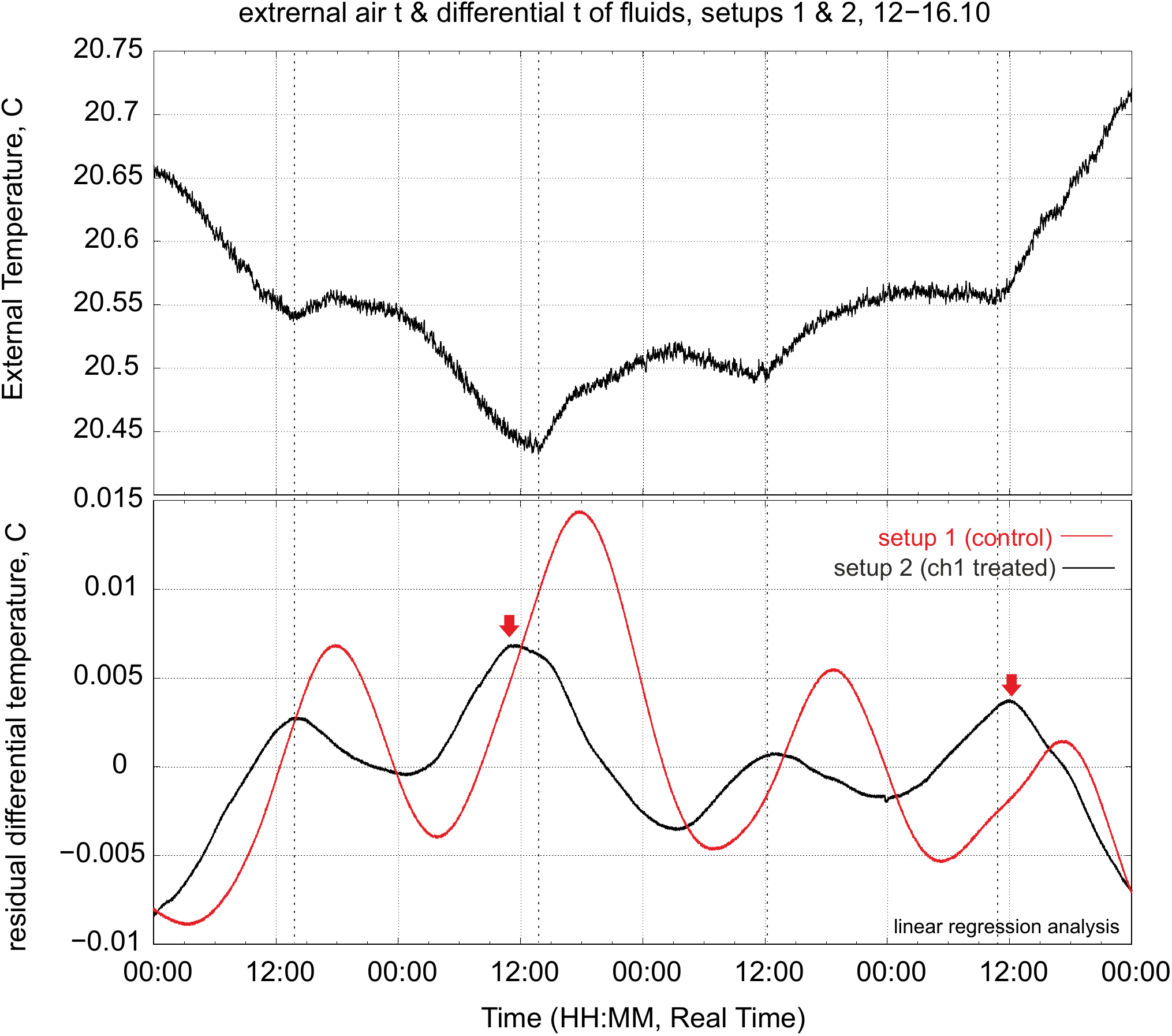}}
\caption{Long-term dynamics of control (with two unexcited samples) and experimental (excited and unexcited samples) calorimeters during 96 hours after excitation: \textbf{(top)} dynamics of environmental temperature in laboratory; \textbf{(bottom)} differential temperature of both passive calorimeters, red arrows show points with leading and lagging differential dynamics. \label{fig:thermalVariations}}
\end{figure}

Second, the system with excited samples reacts differently on thermal variations than the system with two unexcited fluids, see Fig. \ref{fig:thermalVariations}. Here we consider two passive calorimeters: one has two unexcited samples (control system), the second one has excited and unexcited samples (experimental system). Trend of environmental temperature changes trends also in passive calorimeters. Control calorimeter reacts always with the same delay (caused by thermal protection); responses of experimental calorimeter vary across environmental peaks demonstrating leading and lagging differential dynamics.

\begin{figure*}[ht]
\centering
\subfigure[]{\includegraphics[width=0.5\textwidth]{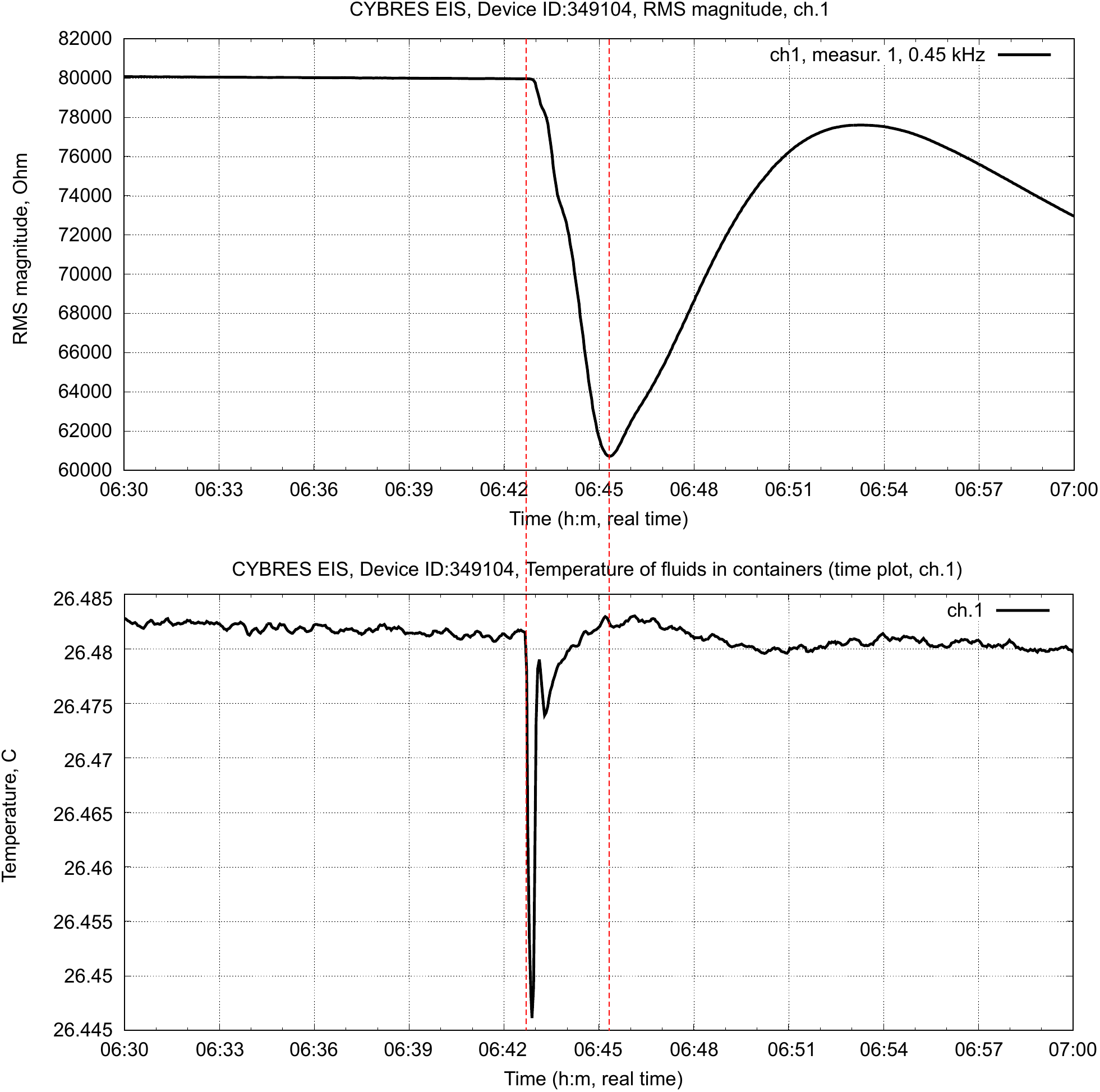}}~
\subfigure[]{\includegraphics[width=0.5\textwidth]{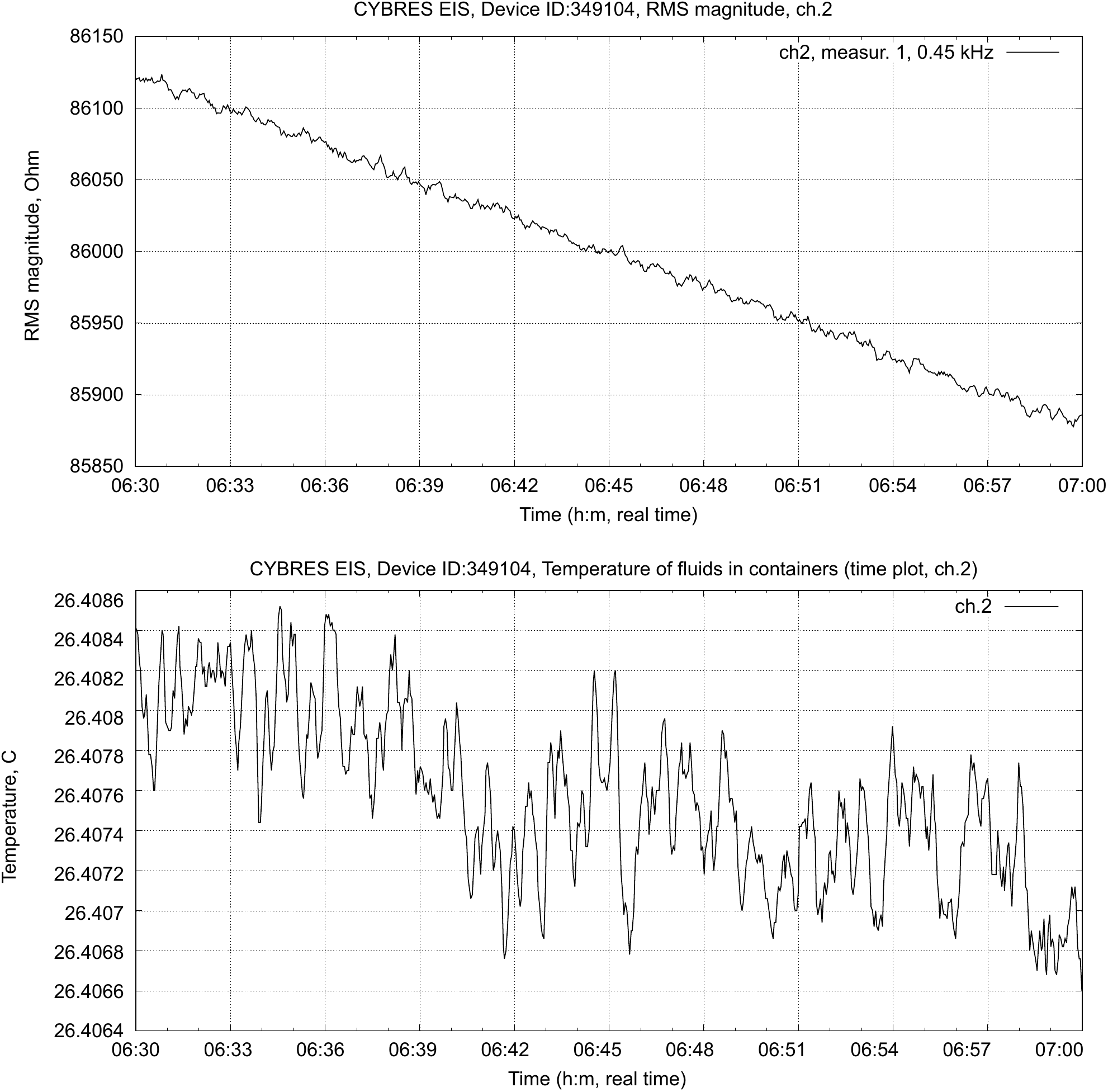}}
\caption{\small An example of spontaneous changes in ionic productivity (inclination of EIS curves) and temperature dynamics on mesoscopic scales: \textbf{(a)} channel 1; \textbf{(b)} channel 2. In channel 1, there is an abrupt change in temperature, which after 1-1.5 minutes begins to affect the impedance. The maximum change in impedance is reached after 3 minutes, when the temperature has already stabilized at the previous level. Further, after 10 minutes, a change in the EIS trend is observed, while the temperature trend is stable. Channel 2 shows no such changes.
\label{fig:spiking_dynamics}}
\end{figure*}

\section{Discussion}
\label{sec:discussion}

Important methodological discussion is related to selection of schemes (\ref{eq:heat4}) or (\ref{eq:heat8}) and two following questions: 1) whether the transient dynamics of scheme (\ref{eq:heat4}) within 60 min after excitation contains reliable data for measuring the heat capacity? 2) whether the spin conversion effects are still present on the scale 90-120 min in the steady-state dynamics of the scheme (\ref{eq:heat8})? We observe about 4.17\%--5.72\% with (\ref{eq:heat4}) and about 2.08\% with (\ref{eq:heat8}), this variation of results can be traced back to changes in the ortho-/para- ratio or to uncontrollable factors during transient dynamics of diathermic measurements. We argue that if the steady-state dynamics of control and experimental samples differ from each other, this should be reflected also in the transient dynamics. Since variation of results in control attempts (StDev=0.0238) is smaller than in experiments (StDev=0.0564) (both attempts have the same uncontrollable factors during transient dynamics), it is expected that the transient dynamics is determined primarily by ortho-/para- ratio.

Thermal measurements are more demanding to the quality of handling and equipment than capillary or electrochemical measurements, thus, the variability of results can be partially explained by slightly different processing of water samples. Repeating arguments from \cite{kernbach2023Pershin}, we would like to point to water interfaces that are considered as candidates for the long-term preservation of non-equilibrium states of spin isomers in the liquid phase \cite{Chai09,Pershin12ICE,Odendahl22}. Refilling of liquids between three different containers during handling essentially changes the structure of interface layers and thus contributes to variability of experimental samples. Considering 30-60 min time scale, discussed e.g. in \cite{Otsuka06,kernbach2023Pershin}, we see essential decreasing of results about 60 minutes after excitation. 

Long-term thermal fluctuations, shown in Figs. \ref{fig:longTermReaction}, \ref{fig:transient_time}, represent another interesting phenomenon. There are several reasons that can explain them: from cosmic particles and anthropogenic factors (e.g. sporadic high-frequency EM emission from mobile phones and WiFi systems) up to technological artifacts appeared only at long-term measurements. For instance, electrochemical measurements \cite{kernbach2022electrochemical,Kernbach17water} with active thermostats demonstrated interesting spiking dynamics of electrochemical impedances and temperature of samples, see Fig. \ref{fig:spiking_dynamics}. The frequency of occurrence of this effect increases with temperature, for example, at $20-25^\circ C$ one spike takes place per 1-3 days, whereas at $>30^\circ C$ 3-4 spikes occur per day. Electrochemical and thermal dynamics does not follow each other in known dependencies \cite{doi:10.1021/ac00140a003}, but are rather fluctuating. This effect hinders long-term electrochemical measurements and forces using thermal profiles with lowering the temperature. Observation with the microscope showed that this effect is not related to microbubbles and possible mechanical stirring of fluids. We can also reject the hypothesis with cosmic particles due to a clear temperature dependency. Characteristic property of these spiking effects is their meso-scale dynamics that differs from chemical kinetics on macro-scales and molecular phenomena on micro-scales (represented e.g. by electrochemical noise). Possible explanation can be expressed in terms of a spin-conversion process affecting both the electrochemical \cite{kernbach2022electrochemical} and thermal properties of aqueous solutions, triggered by spontaneous environmental (thermal, EM or geomagnetic) fluctuations.

\section{Conclusion}

This work demonstrated several short-term and long-term thermal effects appeared after excitation of fluids in mechanical way. Thermal measurements are conducted without additional excitation by electric current (as used in electrochemical measurements), thus, we can exclude from consideration inducted endotermic/exotermic electrochemical reactions. We noted different thermal dynamics of control and experimental attempts in transient states around the first and the second temperature points. Changes of the heat capacity are about 4.17\%--5.72\% in the transient dynamics within 60 min after excitation and about 2.08\% in the steady-state dynamics on the scale 90-120 min after excitation. Short-term and long-term thermal fluctuations are at the level of 10$^{-3}$ $^\circ C$ of relative temperature. Observed thermal effects are reproducible in both active and passive calorimeters, however we noted larger long-term fluctuations in the measurement device with active thermostat that follows the increasing temperature profiles. Lowering the temperature reduces such long-term effects.

As reported in other works \cite{kernbach2022electrochemical, Pershin09Temp, Pershin22}, we do not see arguments to reject the hypothesis about spin conversion taking place in liquid phase, on the contrary we observe multiple side phenomena such as changing a heat capacity, surface tension and capillary effects \cite{kernbach2023Pershin} that must accompany the spin conversion process. Due to low energy required for spin conversion, we can assume that weak energy influx from environment as well as from the thermostat can facilitate an ongoing spin conversion process.

\section{Acknowledgment}

This work is partially supported by EU-H2020 Project 'WATCHPLANT: Smart Biohybrid Phyto-Organisms for Environmental In Situ Monitoring', grant No: 101017899 funded by European Commission.

\small

\end{document}